\documentclass{article}
\usepackage{ijcai25}

\usepackage{times}
\usepackage{soul}
\usepackage{url}
\usepackage[hidelinks]{hyperref}
\usepackage[utf8]{inputenc}
\usepackage[small]{caption}
\usepackage{graphicx}
\usepackage{amsmath}
\usepackage{amsthm}
\usepackage{booktabs}
\usepackage{algorithm}
\usepackage{algorithmic}
\usepackage{framed}
\usepackage{caption}
\usepackage[switch]{lineno}
\usepackage[most]{tcolorbox}
\usepackage{xcolor}
\usepackage{booktabs}
\usepackage[margin=1in]{geometry}
\usepackage{multirow} 
\usepackage{array}  
\usepackage{ragged2e}

\newtcolorbox{promptbox}[1]{
  enhanced,
  colback=teal!5!gray!10,
  colframe=cyan!50!black,
  arc=2mm,
  boxrule=0.5pt,
  title=#1,
  fonttitle=\bfseries\color{white},
  coltitle=green!50!black,
  breakable=false,
  width=\textwidth / 2
}

\title{Evaluation of Small Vision-Language Models on Qualitative Mechanical Problems\thanks{Proceedings of the IJCAI Workshop on ``\textit{38th International Workshop on Qualitative Reasoning}'' (QR 2025).}}

\author{
    Henry Fordjour Ansah$^1$\and 
    Shreya Banerjee$^1$\and 
    Pranish Ghimire$^1$
    \affiliations
    $^1$Louisiana State University of New Orleans
    \emails
    \{hfansah, pghimir3\}@lsuneworleans.edu,
    shreyabbanerjee@gmail.com
}

\begin{document}

\maketitle

\begin{abstract}
    \textit{Qualitative mechanical problem-solving (QMPS)} refers to
  solving qualitative problems from the mechanical domain. Qualitative 
  problems can be solved with minimal discipline-specific information, 
  without any robust quantitative calculation, generally by using qualitative
  reasoning and commonsense knowledge.  QMPS is a vital aspect of human 
  intelligence that allows us to tackle a wide range of tasks, from simple 
  everyday ones such as turning on a tap to complex tasks in highly demanding and
  well-paying jobs in various fields, e.g.,~emergency medicine,
  plumbing, driving, etc.  Employers often use the \textit{Bennett
    Mechanical Comprehension Test (BMCT)} to evaluate job candidates'
  ability to solve such problems. In this work, we assess two state-of-the-art multimodal models, Gemma-3 and Qwen-VL, on their ability to interpret mechanical problem images by eliciting a step-by-step \textit{chain of thought (CoT)} and a final answer. Each image inherently encodes ground-truth qualitative facts, such as contact points in gears, support relations, and relative weights, which we use to evaluate each model's spatial and commonsense reasoning capabilities. We assess each chain for coherence, completeness, and logical progression to assess each model's thought process, and final answers are compared to verified solutions to measure accuracy.
\end{abstract}

\section{Introduction}

Qualitative problems differ from quantitative problems in that the 
 former can be solved with scant discipline-specific information 
 and without any robust quantitative calculation.
 
 \begin{itemize}
	
    \item \textbf{Quantitative Problem Solving:} Solving a problem
      based on the actual quantity of attributes (value of length,
      volume, distance, etc.) as employed in the particular sciences
      (e.g.\ physics).
    \item \textbf{Qualitative Problem Solving:} Solving a problem
      based on the quality of attributes (comparison of size,
      appearance, symmetry, volume, length, etc.) as employed in
      everyday cognition.
\end{itemize}

\textit{Qualitative mechanical problem-solving (QMPS)} is a method of solving mechanical problems through \textit{qualitative reasoning (QR)} 
\cite{forbus1997qualitative}, \cite{weld2013readings} rather than through robust quantitative calculations requiring equations, theorems, formulae, etc., from mechanics/physics textbooks. This concept is brilliantly captured in Hayes' work \cite{hayes1978naive,hayes1985naive} where qualitative reasoning about fluid behavior is leveraged from human general knowledge rather than physics proper. QMPS is integral to human intelligence, allowing 
individuals to perform everyday tasks (such as hanging a picture frame on a wall) as well as complex professional tasks that pay well in today's economy 
(e.g., in emergency medicine, plumbing, and the use of hydraulic machinery), using commonsense knowledge without the need for discipline-specific 
information or rigorous calculations. This skill is lacking in current artificial intelligence (AI) and robotics \cite{banerjee2023qualitative}, marking a significant gap in achieving 
artificial-general intelligence (AGI) in machines.

In this work, we conduct a detailed evaluation of two small vision-language models, Qwen-VL \cite{bai2023qwenvlversatilevisionlanguagemodel} (\textit{Hugging Face ID: Qwen/Qwen2.5-VL-7B-Instruct}) and Gemma-3 \cite{team2025gemma} (\textit{Hugging Face ID: google/gemma-3-4b-it}), on a single A100-40GB GPU, focusing on their ability to solve qualitative mechanical problems. This assessment moves beyond simple answer accuracy by also critically examining the models' underlying thought processes (chain-of-thought). By analyzing both the correctness of their solutions and the coherence and validity of their reasoning, we aim to provide a nuanced understanding of their performance, identify common error patterns, and highlight their respective strengths and weaknesses in this challenging domain.

The following section further describes QMPS, and our motivation for this work. Next, we illustrate our approach and subsequently analyze the results with detailed case studies, followed by discussion and future work.

\section{Motivation and Background}
The motivation for this work stems from the increasing need to understand and benchmark the reasoning capabilities of small vision-language models (SVLMs), such as Gemma and Qwen-VL, particularly for qualitative problem-solving in specialized domains. While large language models such as GPT-4 \cite{achiam2023gpt} have demonstrated impressive feats in several domains, smaller, more efficient models are becoming crucial for broader accessibility and deployment in real-time environments. Evaluating these SVLMs on tasks that require not just language understanding or visual perception in isolation, but the integrated reasoning about mechanical concepts presented visually and described textually, is vital for agents to solve QMPS challenges in the real world at the level of human intelligence, and perhaps beyond.

Under the rubric of \textit{Psychometric AI} \cite{bringsjord2003psychoai} that rigorizes the goals of AI to corresponding human-level tests of cognitive power, for 
QMPS by AI, we focus on problems from the \textit{Bennett Mechanical 
Comprehension Tests (BMCT-II)} \cite{bennett1969bennett}, which we call \textit{BMCT or Bennett}, 
in tandem. The BMCT assesses a human's ability to 
understand mechanical concepts and apply them for problem-solving. As such, it has long been used in the real world by many 
companies, across several different industries, as a recruitment/training tool to evaluate job candidates for roles that require 
innate mechanical acumen. BMCT problems are multiple-choice questions (MCQ) consisting of one or more diagrams with labels, markings and textual information, 
a separate or combined story and question (= stem), with a number of options ranging from 3-5. Challenges of solving such 
problems include representing visual-conceptual relationships, spatial attributes, labels and annotations from the diagrams.

The specific focus on BMCT/Bennett problems was first followed
by \cite{klenk2011usingBMCT}, who reported an analogical approach, whereas recently \cite{banerjee2022qualitative}
showed a formal logicist approach to the Bennett or BMCT
problems. We have evaluated the strengths and weaknesses of a purely statistical approach via vision-language models.

\section{Methodology} \label{sec:experiments}
\subsection{Dataset or Test Cases}
From iPREP BMCT series \cite{iPREP_bmctseries}, we curated 30 benchmark problems from across four scenario types: hydraulics (7), Gears and Belt Drives (6), Acoustics (5), Force and Mechanics (6), and Heat and Radiation (6) as used in the Bennett test, with each domain having 2 easy-level questions, 2 medium-level questions and 2 hard-level questions, except the hydraulics domain, which has 3 medium-level questions.
Each instance of a problem is a graphical depiction of a type of qualitative mechanical problem that encodes all necessary visual cues, such as fluid levels, force vectors, or gear configurations, accompanied by a question and a fixed set of answer choices. We embed each instance into a template with a system prompt shown below, to elicit a chain-of-thought reasoning from the models in a suitable format.\\
\noindent
\textit{SYSTEM PROMPT: Write a response that appropriately completes the request.
Before answering, think carefully about the question and create a step-by-step chain of thoughts
to ensure a logical and accurate response. Put the step-by-step chain of thoughts in a think tag (\texttt{<think>...</think>}) and put your 
final answer which should be from the options provided only in an answer tag (\texttt{<answer>...</answer>}).}

\subsection{Categories of Failure and Success Cases}
In order to understand the sources of errors and unexpected successes in our multimodal reasoning pipeline, we analyze four prototypical cases across the 30 BMCT benchmark problems. 

\subsubsection{Case 1: Flawed Reasoning Leading to Wrong Answer}
In this scenario, the model’s chain of thought (CoT) explicitly mentions all relevant visual facts (high coverage) and the final answer is incorrect. Investigation reveals that the error arises from one or more of the following:  
\begin{itemize}
  \item \emph{Incorrect Assumptions}: The model infers properties not supported by the image (e.g., assuming unequal lever arms when both arms are equal length).  
  \item \emph{Misapplication of Physical Concepts}: Fundamental qualitative laws—such as force equilibrium or torque balance—are applied incorrectly (for example, adding forces when the correct operation is to compare magnitudes qualitatively).  
  \item \emph{Logical Fallacies}: Steps in the CoT contain internal contradictions or invalid deductions (e.g., inferring that a large object must be less dense than a small one because it floats without considering the average density relative to the fluid).  
\end{itemize}
We catalog these error types and compute their frequencies. For each error instance, we also record the CoT quality score to examine whether low coherence or logical progression correlates with flawed concept application.

\subsubsection{Case 2: Omission of Crucial Visual Information - Poor Spatial Reasoning}
Here, the model’s final answer is incorrect primarily because the CoT fails to mention one or more essential facts encoded in the image. Typical omissions include:  
\begin{itemize}
  \item \emph{Missing Contact Points}: The model does not register which object is in contact with which support or surface, such as in the case of multiple gear connections.  
  \item \emph{Ignored Force Vectors or Pressure Levels}: In hydraulics problems, differential fluid heights are overlooked; in gear problems, gear tooth counts or meshing relationships are omitted.  
  \item \emph{Overlooked Geometric Relations}: Relative distances or angles (e.g., incline angle, lever arm lengths) are not referenced.  
\end{itemize}

\subsubsection{Case 3: Correct Answer Despite Flawed Reasoning}
Occasionally, the model arrives at the correct final answer even though the CoT exhibits clear reasoning flaws. These cases fall into two subtypes:  
\begin{itemize}
  \item \emph{Lucky Guessing}: The model hallucinates intermediate facts or applies concepts incorrectly but nonetheless arrives at the correct answer by chance.  
  \item \emph{Partial Reasoning Suffices}: Despite an incomplete or partially incorrect chain, the model captures a minimal set of essential facts correctly, enabling a correct qualitative judgment.  
\end{itemize}
\subsubsection{Case 4: Correct Answer with Flawless Reasoning}

We will also analyze a case where both models not only produce the correct answer but also display coherent, logically sound chains of thought.

\subsection{Results and Analysis}
Table \ref{tab:booktabs} provides an overview of both models' performance across all 30 problems, reporting their overall accuracy as well as the percentage of flawless versus flawed reasoning and failures due to poor spatial reasoning. Below, we further delve into each category with a demonstration, exploring how reasoning quality and spatial understanding contribute to overall success or failure.
\begin{table*}[htbp]
\centering
\caption{Summary of Performance of Qwen-VL and Gemma on 30 BMCT Problems}
\label{tab:booktabs}
    \begin{tabular}{|l|c|c|c|c|}
        \hline
        \multicolumn{1}{|c|}{} & \multicolumn{2}{c|}{Accurate} & \multicolumn{2}{c|}{Inaccurate} \\
        \hline
        \multirow{2}{*}{Gemma} & \multicolumn{2}{c|}{36.67\%} & \multicolumn{2}{c|}{63.33\%} \\
        \cline{2-5}
         & Correct Reasoning & Incorrect Reasoning & Flawed Reasoning & Missed Visual Cue \\
         & 30\% & 6.67\% & 40.00\% & 23.33\% \\
        \hline
        \multirow{2}{*}{Qwen} & \multicolumn{2}{c|}{46.67\%} & \multicolumn{2}{c|}{53.33\%} \\
        \cline{2-5}
         & Correct Reasoning & Incorrect Reasoning & Flawed Reasoning & Missed Visual Cue \\
         & 40.00\% & 6.67\% & 33.33\% & 20.00\% \\
        \hline
    \end{tabular}
\end{table*}
\subsubsection{Case 1: Flawed Reasoning Leading to Wrong Answer}
\begin{figure}[h!]
    \centering
    \includegraphics[width=0.55\textwidth]{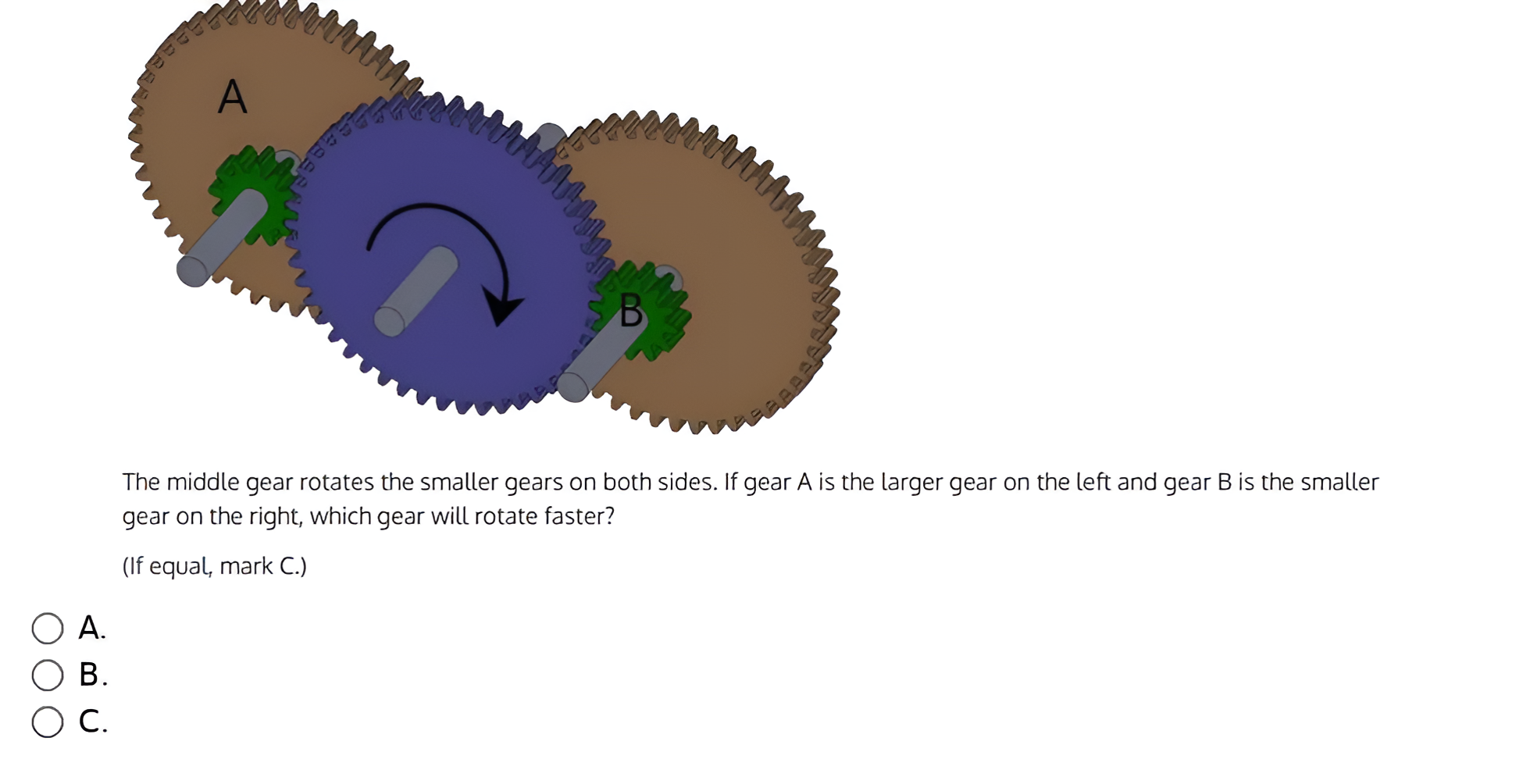}
    \caption{Sample qualitative mechanical reasoning problem from BMCT in the gear domain}
    \label{fig:gear-sample}
\end{figure}
Figure~\ref{fig:gear-sample} shows a peculiar case where both models provide an incorrect answer due to flawed reasoning. In this problem, a central driving gear (purple) meshes with two smaller gears: one labeled B (right) and the other rigidly fixed to a larger gear labeled A (left). Both meshes occur at the same tangential (rim) speed. The question asks: given that the middle gear drives both sides equally, which gear, A or B, will end up rotating faster? 

\subsubsection*{Correct Reasoning Trace}

As we can see, the middle gear directly meshes with the smaller gears on both sides. We know from the basic physics principle that the tangential speed of meshed gears is identical. Since both smaller gears mesh with the middle gear, both would have the same tangential speed. Both are of the same size (radius); hence, the angular speed is also the same.
\footnote{For rigorous treatment, we might recall the fundamental relation:
\[
  v = r \cdot \omega,
\]
where \(v\) is the tangential (rim) speed, \(r\) is the gear’s radius, and \(\omega\) is its angular (rotational) speed.
Because both the tangential speed \(v\) and the radius \(r\) are the same for the two meshed gears, their angular speeds \(\omega\) must also be the same.} Consequently, both smaller gears rotate their shafts at identical rates. Since each outer wheel (A and B) is rigidly fixed to its respective pinion, they too turn at the same angular velocity. Hence, gear A and gear B rotate at the same speed.

Both Gemma and Qwen-VL arrive at the same (incorrect) conclusion: “Gear B rotates faster than Gear A”, and their chains of thought portray distinct logical fallacies when compared to the correct reasoning.

\begin{figure}[!h]
    \centering
    \begin{promptbox}{Case 1 - CoT Illustration - Gemma}

\colorbox{teal!25}{%
  \parbox{\dimexpr\linewidth-2\fboxsep}{The question describes a gear system where a middle gear rotates smaller gears on both sides. Gear A is larger on the left and Gear B is smaller on the right. 
  }%
}
    
We need to determine which gear will rotate faster.
\colorbox{red!25}{%
  \parbox{\dimexpr\linewidth-2\fboxsep}{The relationship between the speed of a driving gear and a driven gear in a gear system is inversely proportional: if the driving gear rotates faster, the driven gear rotates slower, and vice versa.
  }%
}
 The number of teeth on the gears dictates this speed ratio. A larger gear (with more teeth) will cause the meshed smaller gear to spin more times per single rotation of the larger gear.

Because Gear A is larger than Gear B, when Gear A rotates it drives Gear B more quickly (since B must complete more rotations to match the same rim distance). Therefore, Gear B will rotate faster than Gear A.
    \end{promptbox}
    \caption{Gemma’s (incorrect) CoT for gear‐speed problem.}
    \label{fig:case-baseline-challenge}
\end{figure}
Gemma’s reasoning begins correctly by noting that the middle gear meshes with both side gears and that this mesh enforces identical tangential velocities at each contact point (highlighted in green). Nevertheless, it subsequently invokes the standard inverse proportionality between driving‐gear and driven‐gear speeds, and the associated teeth‐count relationship, as if Gear A directly drives Gear B (highlighted in red). This rule, however, is inapplicable to the present configuration, in which both side gears are independently driven by the central gear. By misapplying this driver–driven proportionality and treating the larger left wheel (A) as the “driver” with more teeth, Gemma deduces that the smaller right wheel (B) must rotate more rapidly.

\begin{figure}[!h]
    \centering
    \begin{promptbox}{Case 1 - CoT Illustration - Qwen-VL}
To determine which gear will rotate faster, we need to consider the relationship between the gears and their sizes. Here's the step-by-step reasoning:

\colorbox{teal!25}{%
  \parbox{\dimexpr\linewidth-2\fboxsep}{\textbf{1. Identify the gears}: Gear A is the larger gear on the left, and Gear B is the smaller gear on the right.
  }%
} 
\\
\textbf{2. Understand the rotation}: The middle gear rotates both smaller gears. Since the middle gear is the same size for both smaller gears, it means that the speed at which the middle gear rotates will be the same for both smaller gears.
\\
\colorbox{red!25}{%
  \parbox{\dimexpr\linewidth-2\fboxsep}{\textbf{3. Determine the speed}: Because the middle gear is the same size for both smaller gears, the smaller gear on the right (Gear B) will rotate faster than the smaller gear on the left because it has fewer teeth and thus a higher rotational speed for the same linear distance traveled by the middle gear.
  }%
} 

Given this analysis, Gear B will rotate faster.
    \end{promptbox}
    \caption{Qwen-VL's (incorrect) CoT for gear‐speed problem.}
    \label{fig:case-baseline-challenge}
\end{figure}

Qwen-VL correctly observes that the central gear meshes with two identical smaller gears, thereby imposing the same tangential velocity at each interface. Nevertheless, it then inexplicably asserts that the right-hand gear (B) rotates more rapidly because “it has fewer teeth and thus a higher rotational speed than the smaller left gear.” This reasoning is doubly flawed: the two driven gears seem identical in tooth count, and a gear’s angular velocity under a fixed tangential input depends solely on its own radius rather than on any comparison of tooth counts between them. By conflating tooth count with rotational speed, Qwen-VL introduces an unfounded assumption that leads to the incorrect conclusion that gear B spins faster than gear A.
\\\\
Flaws in the chains of thought produced by both Gemma and Qwen directly lead to their erroneous conclusion that the smaller right‐hand gear spins faster. In each case, the models hallucinate dependencies that simply do not exist in this configuration: Gemma treats Gear A as the “driver” and reverses the teeth‐speed relationship, while Qwen invents a teeth‐count disparity between identical driven gears. These hallucinations of nonexistent facts and misused physical laws compromise the integrity of every subsequent inference, ultimately yielding an inaccurate final answer, pointing towards the crucial need for verifiable proof for their answers.


\subsubsection{Case 2: Omission of Crucial Visual Information - Poor Spatial Reasoning}
\begin{figure}[h!]
    \centering
    \includegraphics[width=0.55\textwidth]{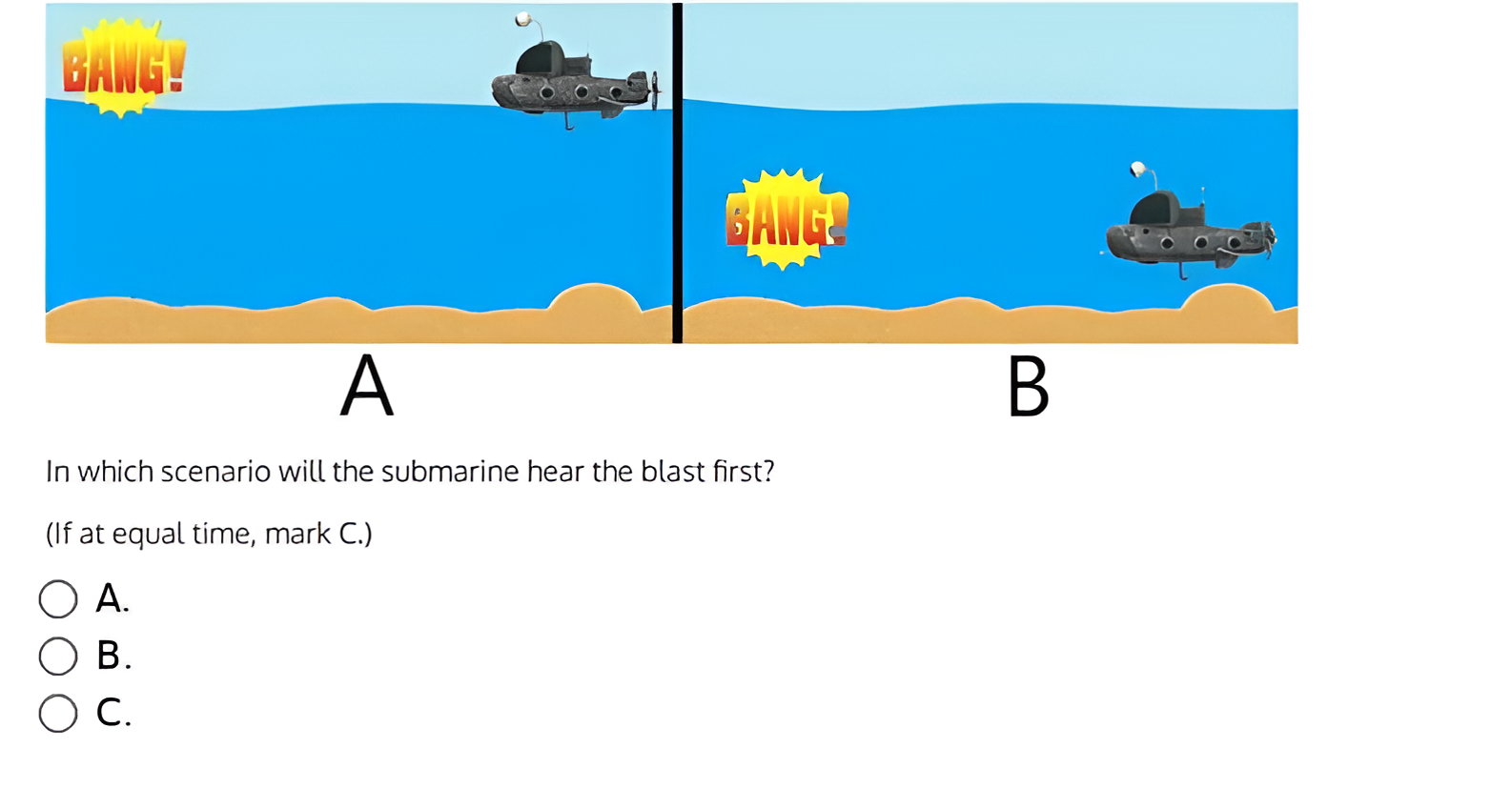}
    \caption{Sample qualitative mechanical reasoning problem from BMCT in the acoustics domain}
    \label{fig:acoustics-sample}
\end{figure}
This problem illustrates a case in which both models fail to answer correctly because they focus on the wrong visual cues. It presents two scenarios for a submarine detecting a blast: in Scenario A the explosion occurs in air and the sound must travel entirely through air before reaching the submarine which is on the surface of the water, whereas in Scenario B the blast originates in the water and propagates through water directly to the submarine which is also underwater. The task is to decide which scenario yields the earliest detection of the sound.
\\\\
\textbf{Correct Reasoning Trace}

Sound is a mechanical wave that requires a medium for propagation, and its speed depends on the medium’s properties \footnote{This is backed by the relation,
\[
  v = \sqrt{\frac{B}{\rho}},
\]
where \(B\) is the bulk modulus and \(\rho\) the density.}. In this problem, the submarine must detect the same blast through either air or water. Because water has a much higher density (and bulk modulus) than air, sound travels substantially faster in water, approximately 1,482 m/s versus 343 m/s in air (and even 5,941 m/s in steel). Consequently, when the explosion occurs in the water (Scenario B), the acoustic wave propagates through the denser medium and reaches the submarine more quickly than in Scenario A, where the sound must traverse the slower, less dense air before reaching the submarine. Therefore, the submarine will hear the blast first in Scenario B.

\begin{figure}[!h]
    \centering
    \begin{promptbox}{Case 2 - CoT Illustration - Gemma}

    The question asks in which scenario the submarine will hear the blast first. We need to analyze the two scenarios depicted.

\colorbox{red!25}{%
  \parbox{\dimexpr\linewidth-2\fboxsep}{Scenario A shows the submarine moving towards a blast that’s already happened. The blast occurs first and the submarine is travelling towards it.
  }%
}

\colorbox{red!25}{%
  \parbox{\dimexpr\linewidth-2\fboxsep}{
Scenario B shows the submarine moving away from a blast that’s already happened. The blast occurs first and the submarine is moving away from it.
  }%
}

Since the question asks which scenario the submarine will hear the blast first, we need to consider the time it takes for the sound to travel. Sound travels at a constant speed. The submarine will hear the blast as it approaches it. In scenario A, the submarine is moving towards the blast, so it will hear it sooner. In scenario B, it's moving away from the blast, so it will hear it later.

Therefore, the answer must be A.
    \end{promptbox}
    \caption{Gemma’s (incorrect) CoT for the acoustic problem.}
    \label{fig:case-baseline-challenge}
\end{figure}

Gemma’s CoT begins by misreading the static illustrations as depicting a moving submarine: it interprets “Scenario A” as the vessel racing toward an already occurred blast and “Scenario B” as it fleeing from one, even though there is no such indication that the submarine is moving. By inventing this nonexistent motion cue, Gemma then invokes irrelevant kinematic reasoning, comparing approach versus recession times rather than attending to the true medium‐dependent propagation speeds. This misidentification of the visual scenarios leads it to apply an inappropriate “sound travels at constant speed relative to a moving observer” argument and thus to the wrong conclusion that the submarine hears the blast sooner in A.
\\


Qwen‐VL similarly misreads the illustrations by presuming an apparent disparity in separation between the submarine and the blast in the two scenarios. It treats Scenario A as if the vessel lies closer to the detonation and Scenario B as if it lies farther away, even though both images depict identical horizontal distances. By inventing this spurious “distance” cue, Qwen then invokes an irrelevant causal factor, emphasizing proximity, rather than recognizing that the acoustic medium is the true determinant. This leads it to conclude, without justification, that the closer submarine in A will hear the blast first, thereby overlooking the critical medium‐dependent speed difference and arriving at a false solution.
\begin{figure}[!h]
    \centering
    \begin{promptbox}{Case 2 - CoT Illustration - Qwen-VL}

   The question asks in which scenario the submarine will hear the blast first.
   \colorbox{red!25}{%
  \parbox{\dimexpr\linewidth-2\fboxsep}{ The key factor here is the distance between the submarine and the explosion.
  }%
}

 \colorbox{red!25}{%
  \parbox{\dimexpr\linewidth-2\fboxsep}{- In Scenario A, the submarine is closer to the explosion compared to Scenario B.
- Sound travels faster through water than through air, but the primary consideration here is the physical distance.
  }
}

Since the submarine is closer to the explosion in Scenario A, it will hear the blast first.

Final Answer: A.
    \end{promptbox}
    \caption{Qwen-VL’s (incorrect) CoT for the acoustic problem.}
    \label{fig:case-baseline-challenge}
\end{figure}

By eliciting explicit chains of thought from these multimodal models, we gain some level of insight into which visual features they attend to and how they integrate those cues into their qualitative reasoning. Instead of treating the models as black boxes whose final response merely reflects internal knowledge, the step-by-step rationales reveal the precise assumptions they derive from each image. 

\subsubsection{Case 3: Correct Answer Despite Flawed Reasoning}
\begin{figure}[h!]
    \centering
    \includegraphics[width=0.55\textwidth]{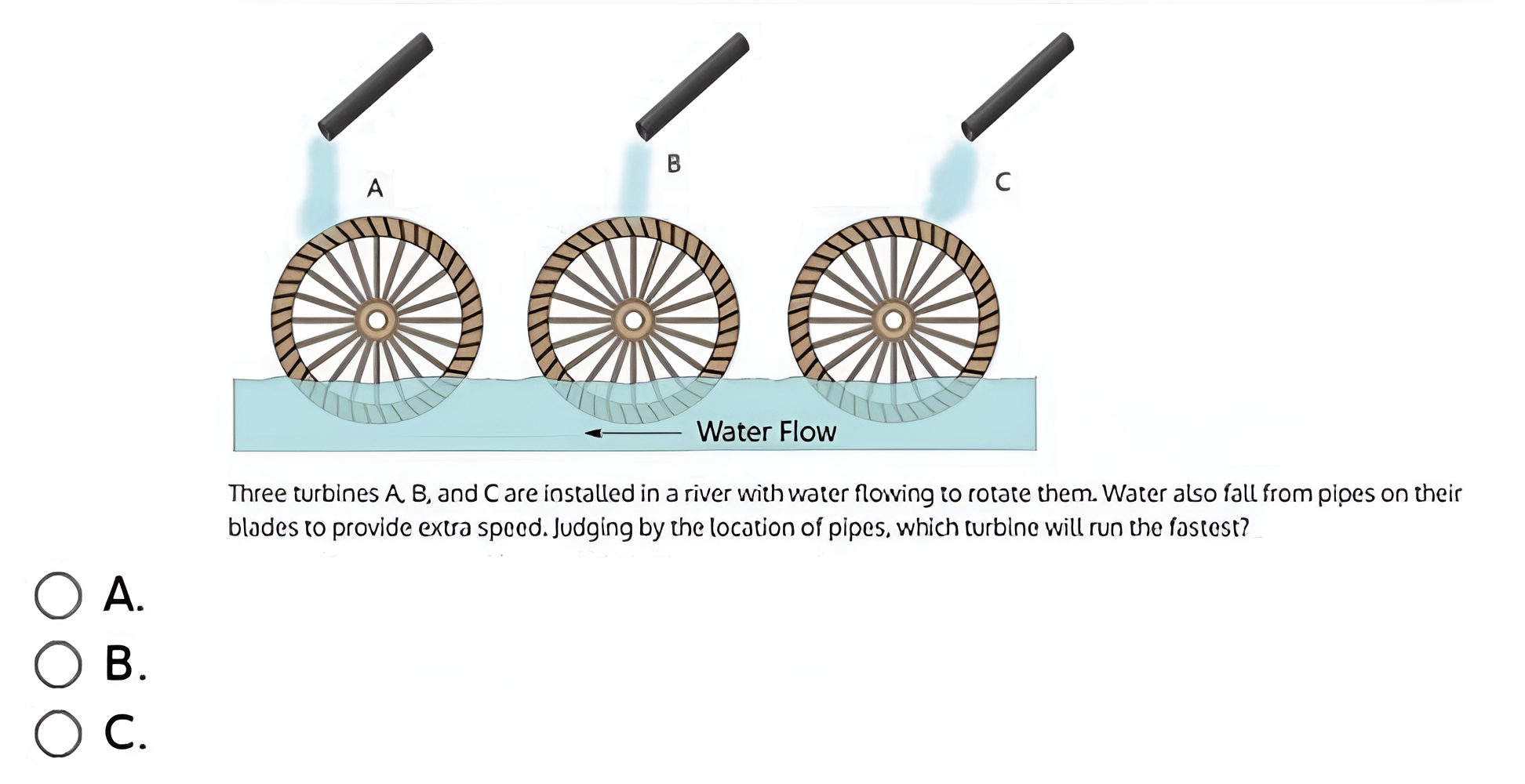}
    \caption{Sample qualitative mechanical reasoning problem from BMCT in the hydraulics domain}
    \label{fig:turbine-sample}
\end{figure}

In this unusual case where the Qwen model luckily arrives at a correct answer, the problem presents a three-turbine setup in which each wheel receives the same water current but also a jet of water from a pipe positioned at different points along its blades. Turbine A’s pipe feeds water at the left side of its blades, Turbine B’s pipe is directly over its pivot, and Turbine C’s pipe delivers water at the right side of its pivot. Although a rigorous torque analysis would show that only Turbine C gains a reinforcing moment from the jet, Qwen instead simplifies the problem to the idea that the speed of the jet water is dependent on where it falls on the blades and that is what determines which blade rotates faster. By focusing solely on water velocity rather than the resultant moment, it nevertheless, and purely by chance, identifies Turbine C (correct answer) as the fastest, despite wrong assumptions.
\begin{figure}[!h]
    \centering
    \begin{promptbox}{Case 3 - CoT Illustration - Qwen}

    To determine which turbine will run the fastest, we need to consider the amount of water flow and the additional speed provided by the pipes on each blade.

 \colorbox{red!25}{%
  \parbox{\dimexpr\linewidth-2\fboxsep}{1. Turbine A: The pipe is located at the top of the blade, which means it will provide the least additional speed because the water will have less time to accelerate before hitting the blade.
  }
}

\colorbox{red!25}{%
  \parbox{\dimexpr\linewidth-2\fboxsep}{2. Turbine B: The pipe is located in the middle of the blade, providing more additional speed than Turbine A but less than Turbine C.
  }
}

\colorbox{red!25}{%
  \parbox{\dimexpr\linewidth-2\fboxsep}{3. Turbine C: The pipe is located at the bottom of the blade, which means it will provide the most additional speed because the water will have more time to accelerate before hitting the blade.
  }
}

Given this information, Turbine C will have the highest additional speed due to the position of its pipe, making it the fastest.
    \end{promptbox}
    \caption{Qwen-VL’s (incorrect) CoT for the hydraulics problem.}
    \label{fig:case-baseline-challenge}
\end{figure}

\textbf{Correct Reasoning Trace}

In the three‐turbine arrangement, each wheel experiences a constant clockwise moment from the river flow, but the falling‐water jet’s contribution depends critically on its position relative to the pivot. For Turbine A, the jet strikes ahead of the pivot, producing a counterclockwise moment that opposes the river’s torque. For Turbine B, the jet impacts directly above the pivot and thus generates no additional moment. In contrast, Turbine C’s jet falls behind the pivot, adding a clockwise moment that reinforces the river‐driven torque. Summing these effects shows that only Turbine C gains extra driving moment, making it the fastest runner.

Such “lucky‐hit” cases where a model’s flawed internal logic nonetheless produces a correct answer highlight both the brittleness and unpredictability of language models in qualitative reasoning.
\\\\
\textbf{Case 4: Accurate Answer and Reasoning}
\begin{figure}[h!]
    \centering
    \includegraphics[width=0.55\textwidth]{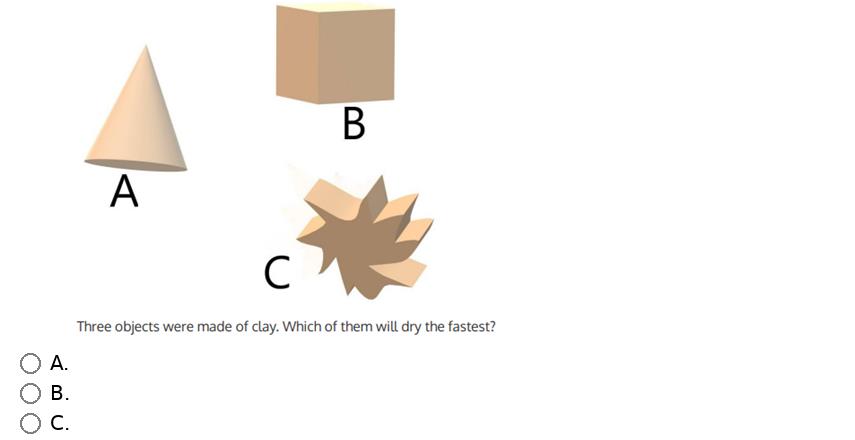}
    \caption{Sample qualitative mechanical reasoning problem from BMCT in the heat-and-radiation domain}
    \label{fig:clay-sample}
\end{figure}
\\
In a fourth evaluation scenario, we examine a problem in which both Gemma and Qwen-VL not only produce the correct answer but also generate coherent, logically sound chains of thought. Here, the cues are unambiguous and the underlying geometric principles are straightforward to apply. Both models accurately extract the necessary details, construct step‐by‐step reasoning that mirrors the ideal solution, and converge on the correct conclusion.

\begin{figure}[!h]
    \centering
    \begin{promptbox}{Case 4 - CoT Illustration - Gemma}
    The question asks which of the three objects made of clay will dry the fastest. Clay dries when water evaporates from its surface. The surface area exposed to the air influences the rate of evaporation. Objects with larger surface areas exposed to air will dry faster.

Let's analyze each object:
\colorbox{teal!25}{%
  \parbox{\dimexpr\linewidth-2\fboxsep}{* A (Cone): The cone has a relatively small surface area compared to its volume. A large portion of the clay is inside the cone and not exposed to the air.
  }
}
\colorbox{teal!25}{%
  \parbox{\dimexpr\linewidth-2\fboxsep}{* B (Cube): The cube has six faces. Each face has a relatively large surface area.
  }
}
\colorbox{teal!25}{%
  \parbox{\dimexpr\linewidth-2\fboxsep}{* C (Fragment): The fragment has a very large surface area due to its many edges and corners.
  }
}
Comparing the three, the fragment (C) has the largest surface area exposed to the air. Therefore, it will dry the fastest.
    \end{promptbox}
    \caption{Gemma’s (correct) CoT for heat-radiation problem.}
    \label{fig:case-baseline-challenge}
\end{figure}

\begin{figure}[!h]
    \centering
    \begin{promptbox}{Case 4 - CoT Illustration - Qwen-VL}

    The question asks which of the three clay objects will dry the fastest. The drying rate of clay objects is influenced by their surface area-to-volume ratio. Objects with a higher surface area-to-volume ratio will dry faster because more water can evaporate from their surfaces.

\colorbox{teal!25}{%
  \parbox{\dimexpr\linewidth-2\fboxsep}{- Object A (cone) has a relatively small surface area compared to its volume.
  }
}
\colorbox{teal!25}{%
  \parbox{\dimexpr\linewidth-2\fboxsep}{- Object B (cube) has a moderate surface area-to-volume ratio.
  }
}
\colorbox{teal!25}{%
  \parbox{\dimexpr\linewidth-2\fboxsep}{- Object C (sculpture) has a very high surface area-to-volume ratio due to its many protruding edges and corners.
  }
}

Given this information, the object with the highest surface area-to-volume ratio will dry the fastest.
    \end{promptbox}
    \caption{Qwen-VL’s (correct) CoT for heat-radiation problem.}
    \label{fig:case-baseline-challenge}
\end{figure}

In the handful of problems where both Gemma and Qwen-VL delivered flawless chains of thought and correct answers, the visual clues were unambiguous/easy-to-understand and the governing principles were nearly straightforward. In these ``easy” cases, the models could reliably map pixels to facts and apply familiar rules without missteps, suggesting that when spatial layouts and qualitative laws are both overt and straightforward, even compact vision-language systems perform excellently. However, this success under ideal conditions also underscores their limitations: as soon as the required visual inference becomes subtle or the physics more nuanced, their spatial reasoning can falter and their qualitative deductions go astray.

\section{Discussion, Conclusion, and Future Work}

The evaluation of Gemma and Qwen on visual reasoning tasks reveals distinct performance profiles and common challenges faced by current language models. While both models demonstrate capabilities in achieving accurate answers with flawless reasoning to a certain extent (Gemma at 30\% and Qwen at 40\%), a considerable number of errors highlight persistent issues in their reasoning.

A significant portion of errors for both models stems from either flawed reasoning or poor spatial reasoning. The category of ``Accurate but Flawed Reasoning," while resulting in a correct final answer, is particularly noteworthy.
It emphasizes that answer-level accuracy alone is insufficient for evaluating true understanding and reliability. 
For real-life purposes, this points towards the importance of underlying justification for LLMs' answers, in the form of verifiable proof/argument.
Flawed reasoning, particularly when it involves the misapplication or hallucination of domain-specific knowledge, points to limitations in the model's grounding and factual recall when applied to complex problems. Poor spatial reasoning highlights the ongoing challenge of robust visual understanding, which is foundational for tasks requiring interpretation of diagrams, scenes, or visual data. We discuss how each model fared in all four cases below.
\\\\
\textbf{Accurate and Flawless Reasoning}: Gemma achieved accurate and flawless reasoning in 9 out of 30 cases (30\%), correctly identifying key visual elements, applying appropriate principles, and following a coherent logical process to arrive at the right answer. Qwen outperformed Gemma slightly in this category, demonstrating accurate and flawless reasoning in 12 out of 30 cases (40\%). In these instances, Qwen reliably interpreted the visual information and applied sound, step-by-step logic without misapplying any physical laws.

\textbf{Accurate but Flawed Reasoning}: Both models produced correct final answers despite flaws in their chains of thought in 2 out of 30 cases each (6.67\%). Gemma’s explanations occasionally contained inaccuracies, such as misstatements of physical principles or minor logical missteps, yet arrived at the right answer by coincidence - a dangerous trend per se. Qwen exhibited a similar pattern - partially logical reasoning or subtle misapplications did not prevent it from selecting the correct option.

\textbf{Inaccurate Due to Flawed Reasoning}: Gemma failed due to flawed reasoning in 12 out of 30 cases (40\%). These errors stemmed from incorrect assumptions, hallucinated concepts, or invalid deductive steps leading to an incorrect conclusion. Qwen exhibited this failure mode in 10 out of 30 cases (33.33\%). 

\textbf{Inaccurate due to Poor Spatial Reasoning}: Gemma’s inability to interpret critical visual details led to errors in 7 out of 30 cases (23.33\%). Even when its internal reasoning was logically coherent, misperceiving spatial relationships or missing key image features caused incorrect premises and, therefore, incorrect outcomes. Qwen encountered this issue in 6 out of 30 cases (20\%).
\\

We observe that addressing these limitations in small vision-language models requires a multi-pronged approach focusing on enhancing both perceptual and reasoning capabilities. In the context of a purely statistical approach for qualitative mechanical reasoning from images, which usually contain spatial relations between objects, we recommend the following techniques to enhance model capabilities: 
\begin{itemize}
    \item{Fine-tuning on Visuo-Spatial Tasks:  Developing and utilizing specialized fine-tuning datasets that explicitly target spatial reasoning skills, such as visual question answering (VQA) datasets focused on physics and engineering that require deep spatial understanding}
    \item{Multi-modal Architectures with Enhanced Visual Encoders:   Exploring architectural improvements that allow for richer and more detailed encoding of visual information, potentially using more powerful vision backbones to capture fine-grained spatial details and relationships between visual elements.}
\end{itemize}

To combat the issue with poor/faulty reasoning, which stems from hallucination, we propose the incorporation of domain-specific knowledge through a neuro-symbolic architecture where atomic facts about physical principles can be embodied in the system. This is an area we are pursuing with a hybrid architecture that combines a fine-tuned LLM with automated reasoning systems to generate trustworthy, verifiable results with proof, accounting for spatial and commonsense reasoning, without compromising the system's speed and robustness.

\section{Acknowledgement}
The authors would like to thank the anonymous referees
for their valuable comments on the earlier version of this
paper.

\bibliographystyle{named}
\bibliography{ijcai25}

\end{document}